\documentclass[conference]{IEEEtran}
\IEEEoverridecommandlockouts
\usepackage{cite}
\usepackage{amsmath,amssymb,amsfonts}
\usepackage{algorithmic}
\usepackage{graphicx}
\usepackage{textcomp}
\usepackage{xcolor}
\def\BibTeX{{\rm B\kern-.05em{\sc i\kern-.025em b}\kern-.08em
    T\kern-.1667em\lower.7ex\hbox{E}\kern-.125emX}}
\begin{document}

\title{Towards Learning a Vocabulary of Visual Concepts and Operators using Deep Neural Networks}

\author{\IEEEauthorblockN{1\textsuperscript{st} Sunil Kumar Vengalil}
\IEEEauthorblockA{
\textit{International Institute of Information Technology}\\
Bangalore, India \\
vengalilsunilkumar@gmail.com}

\and

\IEEEauthorblockN{2\textsuperscript{nd} Neelam Sinha}
\IEEEauthorblockA{
\textit{International Institute of Information Technology}\\
Bangalore, India \\
neelam.sinha@iiitb.ac.in}
}

\maketitle

\begin{abstract}
 Deep neural networks have become the default choice for many applications like image and video recognition, segmentation and other image and video related tasks.
    However, a critical  challenge with these models is the lack of explainability.
    This requirement of generating explainable predictions has motivated the research community to perform various analysis on trained models.
    In this study, we analyze the learned feature maps of trained models using MNIST images for achieving more explainable predictions.
    Our study is focused on deriving a set of primitive elements, here called “visual concepts”, that can be used to generate any arbitrary sample from the data generating distribution.
    We derive the primitive elements from the feature maps learned by the model.
    We illustrate the idea by generating visual concepts from a Variational Autoencoder  trained using MNIST images.
    We augment the training data of MNIST dataset by adding about 60,000 new images generated with visual concepts chosen at random.
    With this we were able to reduce the reconstruction loss (mean square error) from an initial value of 120 without augmentation to 60 with augmentation.
    Our approach is a first step towards the final goal of achieving trained deep neural network models whose predictions, features in hidden layers and the learned filters can be well explained.
    Such a model when deployed in production can easily be modified to  adapt to new data, whereas existing deep learning models need a re-training or fine tuning.
    This process again needs a huge number of data samples that are not easy to generate unless the model has good explainability.
\end{abstract}


\section{Introduction}
Ever since the break through success of Alexnet \cite{krizhevsky2012imagenet} for image classification, deep neural networks has been explored and successfully being used for many complex tasks not only in domains like  vision, speech and natural language processing for which they were originally invented for, but also for other domains in industry.
However, one of the major pain points of these models, when used in industry, is the lack of explainability, i.e., they are unable to provide the exact reason for prediction.
One the one hand, researchers build complex ensemble and deep architectures in order to increase the classification accuracy, which results in increased accuracy but the more complex the model is, the more difficult to provide explanations for the predictions.

In this study, we propose a novel method for  augmenting the training set and introducing losses in hidden layers during training so that the predictions, and the hidden layer feature maps that lead to the final predictions,  are more explainable.
Many existing algorithms \cite{ribeiro2016should} \cite{lundberg2017unified} \cite{zhou2016learning} for building explainability into deep learning models are applied after the model is fully trained.
However, our approach differs significantly as we are modifying the loss function and training data for incorporating explainability.
In addition to solving  the primary task of explainability, our approach also benefits from the fact that  the model can be trained with a smaller number of epochs.
This is because all the hidden layers, where the losses are introduced,  start learning the features in parallel, whereas in existing architectures training losses are computed only at the output layer and it takes a larger number of epochs for the initial layers to learn useful features.
Further, since the loss functions can be introduced at any layer directly, the approach will not suffer from vanishing gradient.

Even though our approach can be applied to any deep learning models, we pick a widely used generative model, Variational Autoencoder \cite{kingma2013auto} in order to illustrate our approach.
Our work is motivated by the concept of learning based on visual concepts  introduced by Lake et.al. in 2015 \cite{lake2015human}.
We generate visual concepts, similar to those used in \cite{lake2015human} and use the images of generated concepts for training a variational autoencoder, whereas in \cite{lake2015human} the concepts are used to build a probabilistic prediction framework.

Major contributions of this paper are:
\begin{enumerate}
  \item Propose a mechanism for data augmentations so that the trained model has better explainability at each layer and feature maps in hidden layers.
  \item Propose a mechanism to introduce losses at hidden layers in a hierarchical fashion starting from primitive visual concepts and building up more complex images by applying some predefined operations on these concepts.
\end{enumerate}

\section{Related Work}
Adding explainability to deep neural networks has been one of the key focus areas of the deep learning research community in the last decade.
Explainability techniques can widely be classified as model-agnostic (which can be applied to any model) or model-dependent where the approach depends on the specific prediction algorithms used by the model.
The early works on this includes the use of explanation vectors introduced by David Baehrens et.al. in their paper \cite{baehrens2010explain}.
Local Interpretable Model-Agnostic Explanations (LIME) \cite{ribeiro2016should} is one of the most widely used tools to explain the predictions of any machine learning model.
LIME works by approximating the model behaviour locally by a simple interpretable surrogate model like linear model or decision tree. The approximation is restricted to a small local change in the input. As this technique looks only at the input and output of the model, this technique can be applied to any machine learning models.
However, the main drawback of the approach is that the surrogate model is only a local approximation and fails to capture the global behaviour.
Rabold et. al. in their recent work \cite{rabold2019enriching} builds on top of LIME and suggests an approach where classifiers' decisions can be explained in terms of logic rules. Their approach is unique in the sense that the predictions can be explained using relationships of objects within the image, whereas in most other approaches the decision is based on presence or absence of certain features.

Another famous technique called Layer-wise Relevance Propagation was introduced by Bach et.al. \cite{bach2015pixel} where the classifier decision is  back propagated and a relevance score is computed at each of the layers backwards until the input units are assigned a relevance score.
The work by Wojciech Samek et.al. \cite{samek2016evaluating} provides a quantitative  evaluation of Layer-wise Relevance Propagation and  suggests it as a better model as opposed to sensitivity based methods.

Scott et.al. introduced a unified approach for explainability which combines multiple methods and brings in a new class of additive feature importance measure (called SHapley) \cite{lundberg2017unified}.
Their method, known as SHapley Additive ExPlanations (SHAP) considers all  the interactions between features and provides an average measure of importance for each feature.

Another class of widely used algorithms for  Convolutional Neural Networks  targets at finding and highlighting the  regions in the input image that is responsible for the prediction.
These techniques, originally introduced in paper \cite{zhou2016learning}, generate a heatmap called Class Activation Map that will highlight the most sensitive regions in the input image.
Saliency Map, introduced by Symonyan et.al. in 2014 \cite{simonyan2014deep} is another variant of class activation map computed by optimizing the input image using the gradient of output prediction with respect to the input image.

\section{Proposed Method}
\subsection{Network Architecture}

\begin{figure}[!t]
\centering
\includegraphics[width=0.9\linewidth]{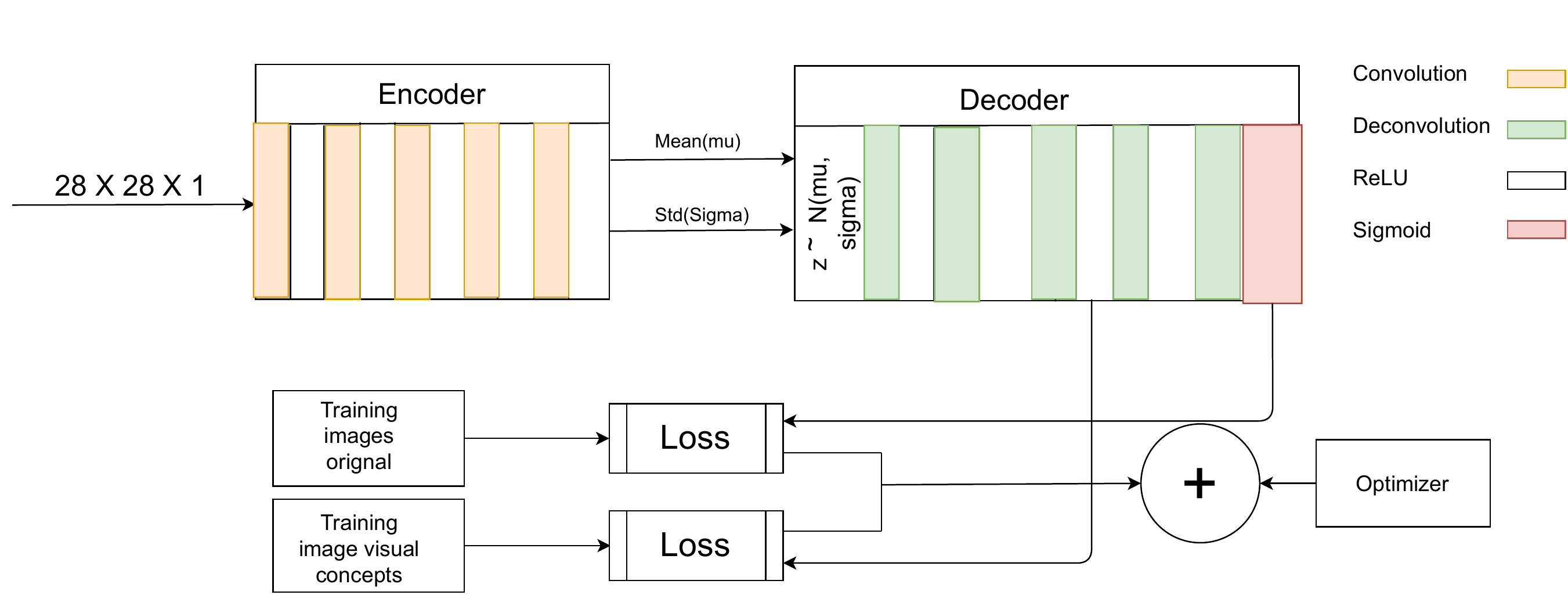}
\vspace{0.2in}

\caption{Architecture of the propsed model. We add an additional loss component at layer 3 of decoder in Variational Autoencoder.}
\label{cnn_arch}
\end{figure}

We use a variational autoencoder (VAE) with five layers of convolutional layers in the encoder and five layers of deconvolution layers in the decoder as shown in Figure \ref{cnn_arch}.
All layers, including the middle layer that generates the latent images, are fully convolutional.
We replaced the middle dense layer of in VAE, with a convolutional layer in order to create a latent image (as opposed to a one dimensional latent vector) that preserves the spatial relationship.
We added an additional loss component at layer 3 of decoder in order to force this layer to learn the premitive visual concepts of the dataset.
The final layer of decoder has sigmoid activation function and we use binary cross entropy loss as the MNIST images can be treated as binary image.

\subsection{Dataset}
We conducted experiments on MNIST dataset by training the VAE with and without concept loss added.
For adding the concept loss, the MNIST training set was first augmented by adding 3000 training images for each of the 18 concepts shown in Figure \ref{visual_concept}.
Some of the similar looking concepts in Figure \ref{visual_concept}, like the horizontal line segments and vertical line segments, were combined into one group to form 18 unique  visual concepts.
The steps for generating concepts are detailed in the sections below.

\subsection{Generating Visual Concepts}
\begin{figure}[!t]
\centering
\includegraphics[width=3.2in]{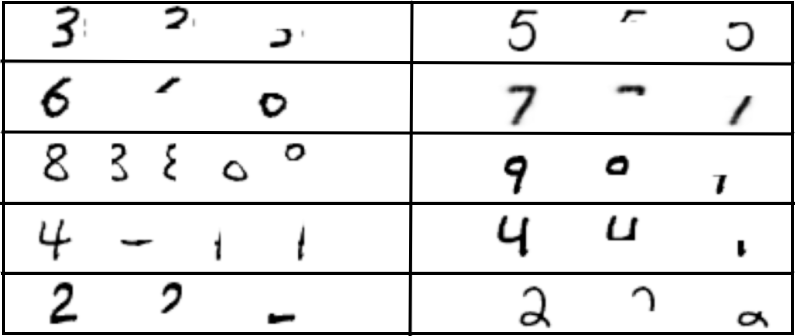}
\vspace{0.1in}

\caption{Visual concepts generated from each of the digits in MNIST. Since there are two different types of images for digits 4 and 2 separate concepts were generated for each image. The width, height and location of these segments were used to define the mean value of a Normal distribution used for generating more samples.}
\label{visual_concept}
\end{figure}

We augment the training set using images of randomly generated segments from MNIST images.
For each of the visual concepts shown in Figure \ref{visual_concept}, 3000 images were generated by randomly sampling the height, width and location of the segment from a normal distribution.
For each concept, the mean value of the height, width and location is kept the same as in Figure \ref{visual_concept}.
The number of visual concepts and their corresponding distributions are hyper parameters that will vary from dataset to dataset.
To generate the segments for each concept, we sampled heights, widths and location of top left corner (represented by two distributions for  $x$ and $y$ coordinates) from the respective concept's distributions and generated slices from randomly chosen training images.
Standard deviation for all distributions were kept as 1 pixel in our experiments.

\subsection{Concept Loss}
We introduce a loss, which we call “concept loss” per  decoder layer and per feature map in the selected hidden layers of the decoder of variational autoencoders (VAE).
Sample visual concepts derived from the training images are used as ground truth for computing the loss for this layer.
For our experiments, we used mean squared error between the feature map and concept ground truth as the loss component.
In our case, the encoder and decoder had 5 hidden layers and the concept loss was applied at the third hidden layer as this is closer to the output layer.
We leave the fourth layer of decoder to learn new derived images from the primitive visual concepts introduced at the third layer.
Mean squared error between original training images and the predicted images is used as loss at the output layer as in usual VAE.

\subsection{Generating Training Samples}

The below steps were followed  to generate additional random samples for each of the  visual concept in Figure \ref{visual_concept}
\begin{enumerate}
\item Create multiple clusters with representative images for each class in the training dataset using below steps
 \begin{enumerate}
     \item Cluster the latent images, learned by VAE,  of all training images into 10 clusters.
      \item Decode the cluster cluster center images using the decoder of VAE. The decoded images are shown in Figure \ref{cluster_center_lv_class}. This forms a set of 10  representative images one for each digit in MNIST dataset.
  \end{enumerate}
\item Choose a random image, either from the training set (with probability  $p$) or from the cluster center image (with probability $1-p$).
 The parameter $p$ can be adjusted to control the diversity of samples generated for each visual concept.
A value of $p$ closer 1 will generate visual concepts from individual training samples most of the time and hence will create a dataset with large variance.
On the other hand, a value close to zero will generate most of the concepts from cluster center images and hence will have less variance. In our experiments, we used $p=0.8$.
\item Get the value of height, width and location of the segment by sampling from the respective distributions.
\item Take a segment from the selected image of the digit with the height, width and location.
\end{enumerate}

Each generated sample is given a label based on the visual concept from which it was generated.

The final training set is generated by combining the original 60,000 training images in MNIST dataset and the 54,000 (300 for each of the 18 concepts) generated concepts to form a training set with 114,000 images.
The labels for the original training samples were kept the same (0-9) as in MNIST dataset and additional 18 labels (10-27) were used for each of the visual concepts.
We used 70\%  of these samples for training and the remaining 30\% were used for validation.

\section{Results and Discussions}
We trained the network for 50 epochs and performed the following clustering operations on the latent vectors and decoder layer 3 feature maps.
\begin{enumerate}
    \item Cluster the latent vectors corresponding to all the training images with label 0-9 (i.e., the training samples for the original MNIST images) into 10 clusters.
    \item Cluster the latent vectors corresponding to all the training images with label 10-27 (i.e., the training samples for the generated visual concept) into 18 clusters.
    \item Cluster the layer 3 feature maps with concept loss added into 18 clusters.
    \item Cluster the layer 3 feature maps without concept loss into18 clusters.
    \end{enumerate}

\begin{figure}[!t]
\centering
\includegraphics[width=3.2in]{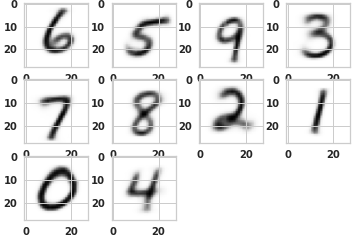}
\vspace{0.1in}

\caption{Decoded Cluster center images of latent vectors corresponding to images from original MNIST classes}
\label{cluster_center_lv_class}
\end{figure}

Cluster centers of latent images were decoded using the decoder of the VAE.
Figure \ref{cluster_center_lv_class} shows the cluster center images corresponding to original images present in the MNIST training set.
As is evident from the figure each cluster corresponds to one of the MNIST class.
The average reconstruction loss for the validation samples reduced from 120 to  60 with the augmented data and added reconstruction loss.

\begin{figure}
\centering
\includegraphics[width=3.2in]{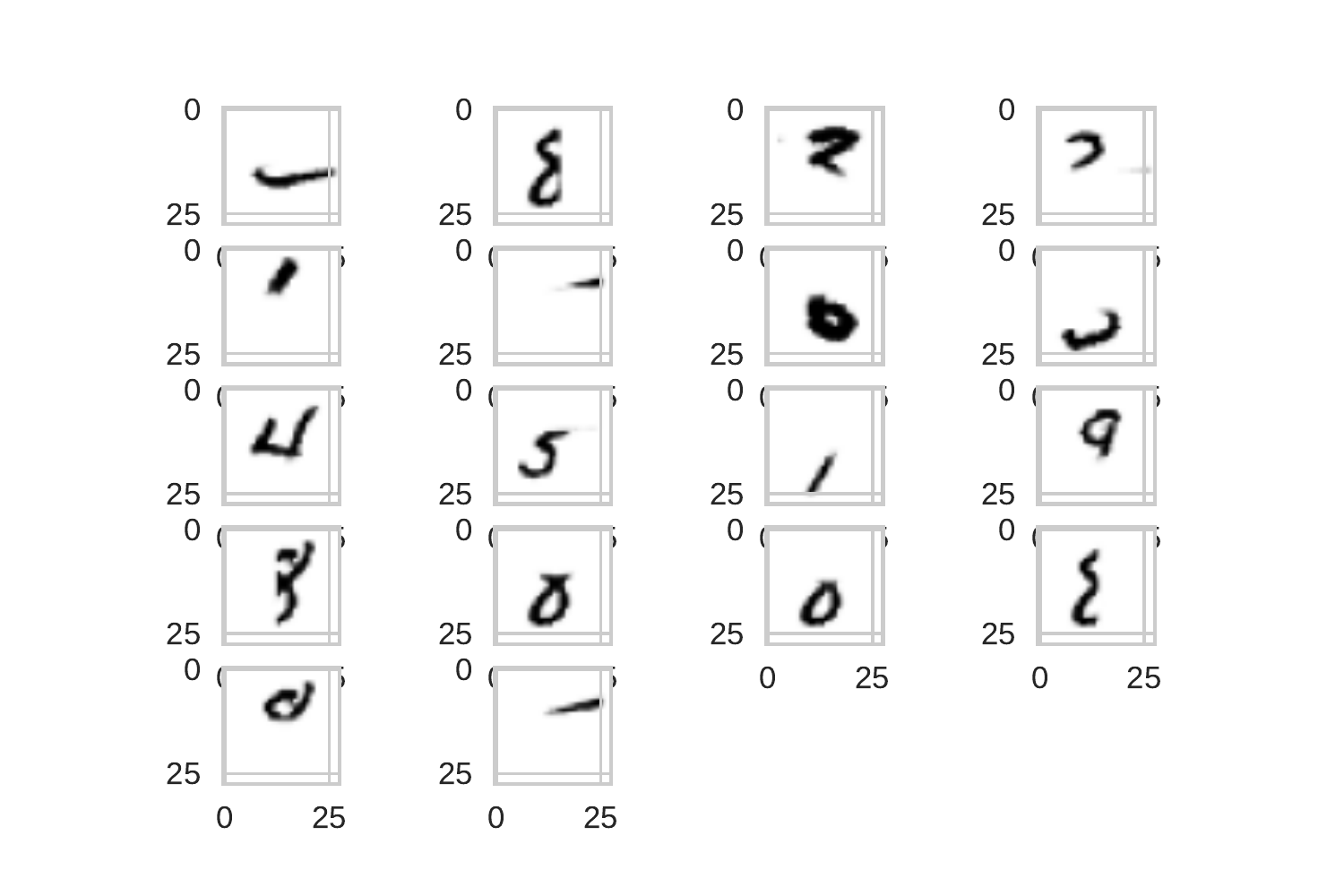}
\vspace{0.1in}

\caption{Cluster centers of the images corresponding to visual concepts. The images were obtained after clustering latent vectors corresponding to images of visual concepts and then decoding the cluster centers. }
\label{cluster_center_lv}
\end{figure}

Figure \ref{cluster_center_lv} shows the cluster centers for the images corresponding to the visual concepts.
As evident from the figure, VAE is able to reconstruct all the generated visual concepts.

\begin{figure}
\centering
\includegraphics[width=3.2in]{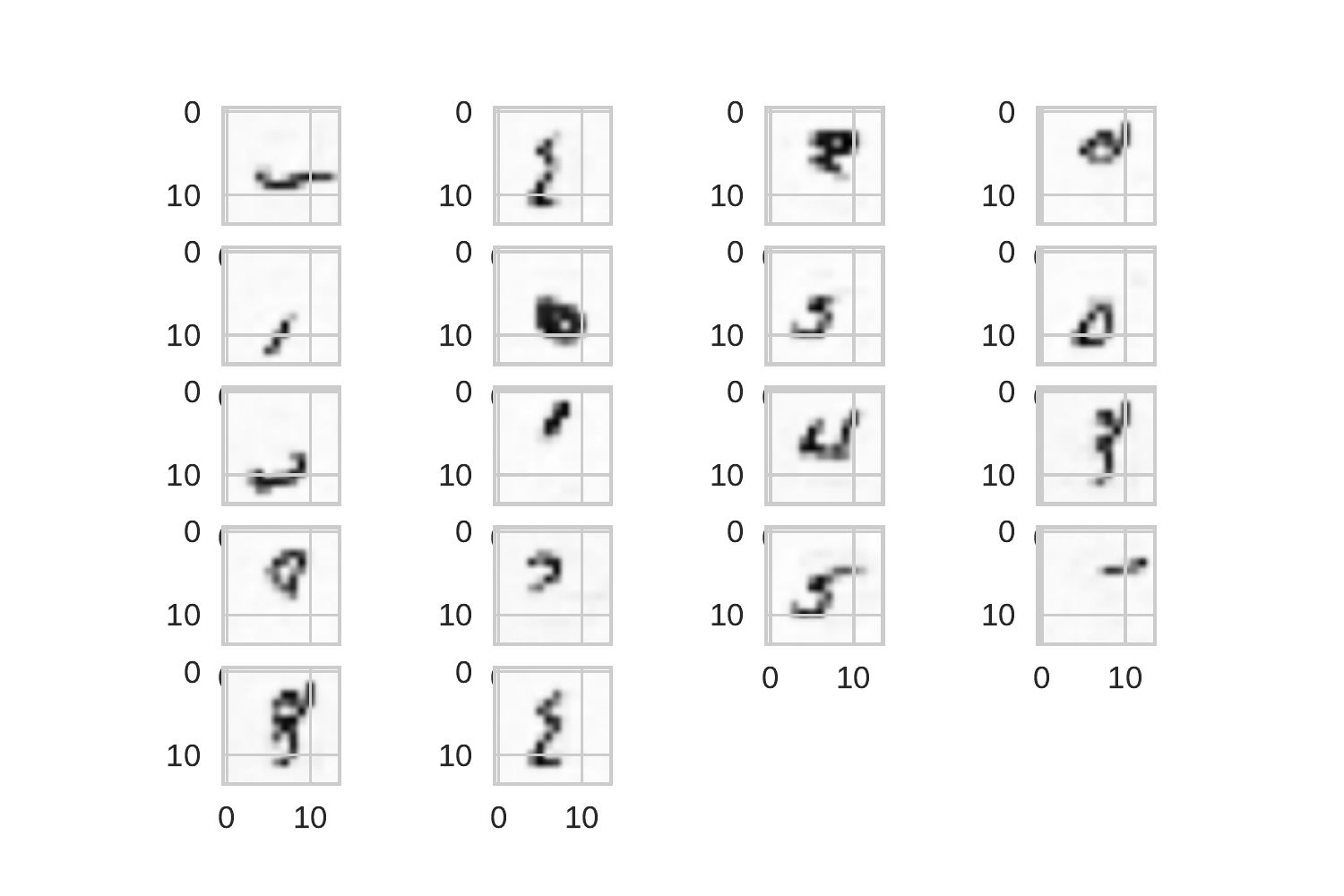}
\vspace{0.1in}

\caption{Cluster center images of layer 3 feature maps with concept loss added}
\label{cluster_center_l3_with_loss}
\end{figure}

\begin{figure}
\centering
\includegraphics[width=3.2in]{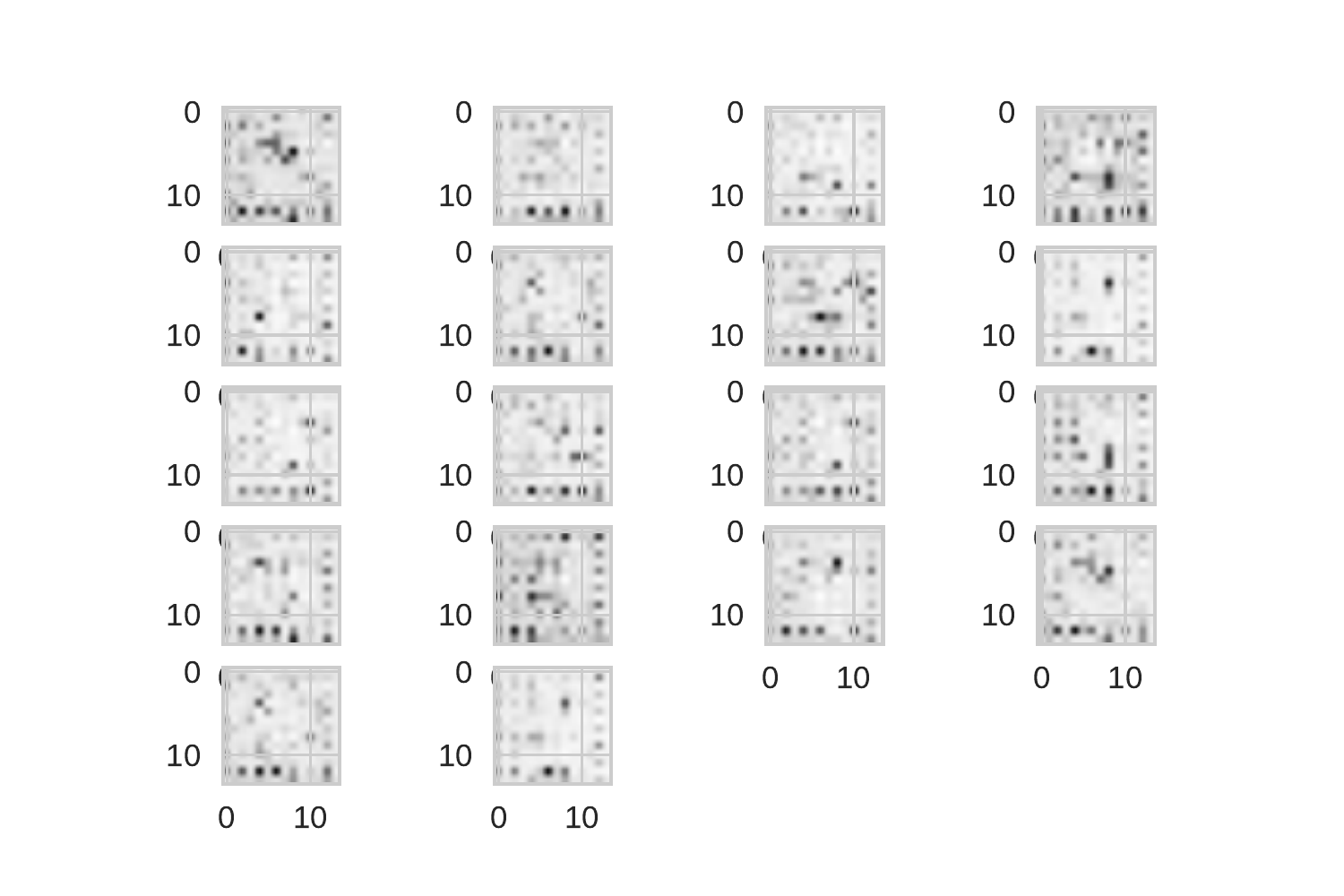}
\vspace{0.1in}

\caption{Cluster center images of layer 3 feature maps without concept loss}
\label{cluster_center_l3}
\end{figure}

Figure \ref{cluster_center_l3_with_loss} and Figure \ref{cluster_center_l3} shows the comparison of feature maps in layer 3 with and without  concept loss added.
As evident from this figure, when the concept loss is added the feature maps in layer 3 is able to learn to reproduce the visual concepts.

%\begin{table}[htbp]
%\caption{Table Type Styles}
%\begin{center}
%\begin{tabular}{|c|c|c|c|}
%\hline
%\textbf{Table}&\multicolumn{3}{|c|}{\textbf{Table Column Head}} \\
%\cline{2-4}
%\textbf{Head} & \textbf{\textit{Table column subhead}}& \textbf{\textit{Subhead}}& \textbf{\textit{Subhead}} \\
%\hline
%copy& More table copy$^{\mathrm{a}}$& &  \\
%\hline
%\multicolumn{4}{l}{$^{\mathrm{a}}$Sample of a Table footnote.}
%\end{tabular}
%\label{tab1}
%\end{center}
%\end{table}

\section{Conclusion}
In this work, we propose a novel method to augment the training data and add an additional loss component in the hidden layer for deep neural networks so that the features in the trained model, and hence the final prediction, is more explainable.
We generate random images of primitive visual concepts, with a predefined distribution, and augment the training set with these images. An additional loss component is added for the generated samples and the loss is directly computed on the feature maps of the hidden layer.
We verified our approach by training a Variational Autoencoder on MNIST images.
We analyzed the feature maps of hidden layers and our analysis shows that the network is able to learn more explainable and semantically meaningful features when the concept loss is added.
Further, we see significant reduction in reconstruction loss from 120, without using concept loss, to a value of 60 with concept loss added.
Our approach is not restricted to VAE and can be applied to any other deep learning model used for  tasks like classification and segmentation.

%\begin{thebibliography}{00}
%\bibitem{b1} G. Eason, B. Noble, and I. N. Sneddon, ``On certain integrals of Lipschitz-Hankel type involving products of Bessel functions,'' Phil. Trans. Roy. Soc. London, vol. A247, pp. 529--551, April 1955.
%\bibitem{b2} J. Clerk Maxwell, A Treatise on Electricity and Magnetism, 3rd ed., vol. 2. Oxford: Clarendon, 1892, pp.68--73.
%\bibitem{b3} I. S. Jacobs and C. P. Bean, ``Fine particles, thin films and exchange anisotropy,'' in Magnetism, vol. III, G. T. Rado and H. Suhl, Eds. New York: Academic, 1963, pp. 271--350.
%\bibitem{b4} K. Elissa, ``Title of paper if known,'' unpublished.
%\bibitem{b5} R. Nicole, ``Title of paper with only first word capitalized,'' J. Name Stand. Abbrev., in press.
%\bibitem{b6} Y. Yorozu, M. Hirano, K. Oka, and Y. Tagawa, ``Electron spectroscopy studies on magneto-optical media and plastic substrate interface,'' IEEE Transl. J. Magn. Japan, vol. 2, pp. 740--741, August 1987 [Digests 9th Annual Conf. Magnetics Japan, p. 301, 1982].
%\bibitem{b7} M. Young, The Technical Writer's Handbook. Mill Valley, CA: University Science, 1989.
%\end{thebibliography}

\bibliographystyle{ieeetr}
\bibliography{ms}

\end{document}